\documentclass[letterpaper, 10 pt, conference]{ieeeconf}  

\IEEEoverridecommandlockouts                              
\usepackage{balance}

\usepackage{xcolor}
\usepackage{comment}
\usepackage{graphics} 
\usepackage{graphicx}
\usepackage{subcaption}
\usepackage{epsfig} 
\usepackage{amsmath} 
\usepackage{amssymb}  
\usepackage{mathtools}
\usepackage{marginnote}
\usepackage{bm}
\usepackage{hyperref}
\usepackage{iitem}
\usepackage{float}
\usepackage{booktabs}

\usepackage{svg}

\newcommand{\block}[1]{\noindent{\textbf{#1}:}}

\newcommand{\Fig}[1]{{Fig.~#1}}
\newcommand{\Table}[1]{{Table~#1}}

\newcommand{\Sec}[1]{{Sec.~#1}}

\newcommand{\cW}{\mathcal{W}}
\newcommand{\cA}{\mathcal{A}}
\newcommand{\cB}{\mathcal{B}}
\newcommand{\cL}{\mathcal{L}}
\newcommand{\cR}{\mathcal{R}}

\newcommand{\be}{\mathbf{e}}
\newcommand{\bx}{\mathbf{x}}
\newcommand{\bu}{\mathbf{u}}
\newcommand{\boldf}{\mathbf{f}}
\newcommand{\bg}{\mathbf{g}}
\newcommand{\bh}{\mathbf{h}}
\newcommand{\bzero}{\mathbf{0}}
\newcommand{\bq}{\mathbf{q}}
\newcommand{\bv}{\mathbf{v}}
\newcommand{\ba}{\mathbf{a}}
\newcommand{\bomg}{\bm{\omega}}
\newcommand{\bR}{\mathbf{R}}
\newcommand{\bp}{\mathbf{p}}
\newcommand{\bW}{\mathbf{W}}
\newcommand{\btau}{\bm{\tau}}
\newcommand{\bw}{\mathbf{w}}
\newcommand{\bJ}{\mathbf{J}}
\newcommand{\bJdot}{\dot{\mathbf{J}}}
\newcommand{\cC}{\mathcal{C}}
\newcommand{\bM}{\mathbf{M}}
\newcommand{\bH}{\mathbf{H}}
\newcommand{\bxi}{\bm{\xi}}
\newcommand{\bphi}{\bm{\phi}}
\newcommand{\bA}{\mathbf{A}}

\newcommand{\WeightedLtwoNormSquare}[2]{\left\lVert #1 \right\rVert_{#2}^2}
\newcommand{\LtwoNormSquare}[1]{\left\lVert #1 \right\rVert_2^2}
\newcommand{\LtwoNorm}[1]{\left\lVert #1 \right\rVert_2}

\newcommand{\txtframe}{\text{frame}}
\newcommand{\human}{\text{human}}
\newcommand{\robot}{\text{robot}}
\newcommand{\quat}{\text{quat}}
\newcommand{\euler}{\text{euler}}
\newcommand{\length}{\text{length}}

\newcommand{\ankle}{\text{ankle}}
\newcommand{\corrected}{\text{corrected}}
\newcommand{\txtroot}{\text{root}}
\newcommand{\EE}{\text{EE}}
\newcommand{\prev}{\text{prev}}
\newcommand{\track}{\text{track}}
\newcommand{\reg}{\text{reg}}
\newcommand{\swing}{\text{swing}}
\newcommand{\contact}{\text{contact}}
\newcommand{\guess}{\text{guess}}
\newcommand{\IK}{\text{IK}}
\newcommand{\KTO}{\text{KTO}}
\newcommand{\ID}{\text{ID}}
\newcommand{\KDTO}{\text{KDTO}}
\newcommand{\mpbpe}{\text{mpbpe}}

\title{\LARGE \bf
SPARK: Skeleton-Parameter Aligned Retargeting on Humanoid Robots with Kinodynamic Trajectory Optimization} 

\author{Anonymous Authors}
\author{ Hanwen Wang$^{1}$, Qiayuan Liao$^{2}$, Bike Zhang$^{2}$, Kunzhao Ren$^{1}$, Koushil Sreenath$^{2}$, Xiaobin Xiong$^{1,3}$ 
\thanks{$^1$Hanwen Wang and Kunzhao Ren are with the University of Wisconsin-Madison. 
        $^2$Qiayuan Liao, Bike Zhang, and Koushil Sreenath are with the University of California, Berkeley.
        $^3$Xiaobin Xiong was with the University of Wisconsin-Madison, and now with the Shanghai Innovation Institute (SII).
        {Corresponding to Koushil Sreenath: \tt\small koushils@berkeley.edu}.
        {Corresponding to Xiaobin Xiong: \tt\small xiaobin.xiong@sii.edu.cn}.}%
}
\begin{document}

\maketitle
\thispagestyle{empty}
\pagestyle{empty}

\begin{abstract}
Human motion provides rich priors for training general-purpose humanoid control policies, but raw demonstrations are often incompatible with a robot’s kinematics and dynamics, limiting their direct use. We present a two-stage pipeline for generating natural and dynamically feasible motion references from task-space human data. First, we convert human motion into a unified robot description format (URDF)-based skeleton representation and calibrate it to the target humanoid’s dimensions. By aligning the underlying skeleton structure rather than heuristically modifying task-space targets, this step significantly reduces inverse kinematics error and tuning effort. Second, we refine the retargeted trajectories through progressive kinodynamic trajectory optimization (TO), solved in three stages: kinematic TO, inverse dynamics, and full kinodynamic TO, each warm-started from the previous solution. The final result yields dynamically consistent state trajectories and joint torque profiles, providing high-quality references for learning-based controllers. Together, skeleton calibration and kinodynamic TO enable the generation of natural, physically consistent motion references across diverse humanoid platforms.
\end{abstract}

\section{Introduction}
Humanoid robots promise to operate in human-centered environments, performing tasks that require rich whole-body coordination, dexterous contact interaction, and agile locomotion. Human motion datasets~\cite{AMASS:ICCV:2019} provide a rich source of coordinated behaviors that can serve as motion priors for training humanoid control policies. Recent retargeting methods~\cite{luo2023perpetual, araujo2025retargeting} have shown that human demonstrations can be converted into natural-looking humanoid motions by mapping human keyframes to robot inverse kinematics (IK) targets, which can then be robustly tracked by controllers trained with reinforcement learning (RL)~\cite{mittal2025isaac, zakka2025mujoco, zakka2026mjlab}. However, most existing retargeting tools model human-robot morphology mismatch using root-link-centered task-space scalings and per-keyframe local offsets~\cite{araujo2025retargeting} that ignore skeletal structure, often requiring substantial IK objective tuning. Moreover, most retargeting pipelines are purely kinematic~\cite{luo2023perpetual, araujo2025retargeting, yang2025omniretarget}, producing dynamically infeasible references that entangle motion tracking with dynamics feasibility recovery in RL.

To address these challenges, we propose a two-stage pipeline that improves both kinematic and dynamic fidelity. First, instead of root-to-keyframe scaling and keyframe local offsets, we generate a human skeletal URDF from task-space motion and calibrate it to the target robot dimensions. Unlike prior human skeleton calibration strategies~\cite{luo2023perpetual} which are often limited to SMPL~\cite{loper2023smpl}, our URDF calibration applies to any human motion format with an implicit skeleton, improving IK quality while reducing tuning efforts.
\begin{figure}[t]
    \centering
    \includegraphics[width=\linewidth]{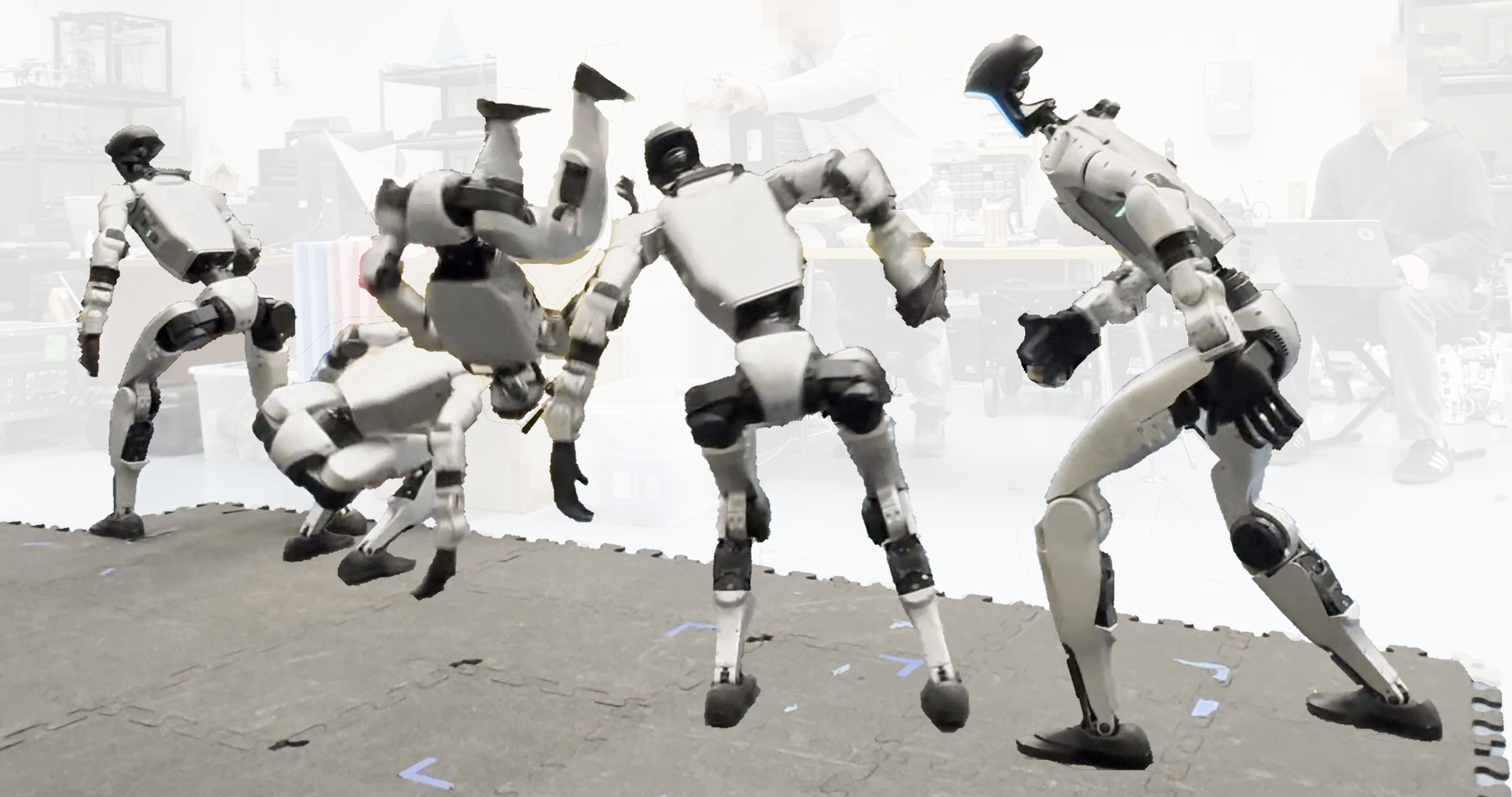}
    \caption{Experimental results of highly dynamic side flip tracking on Unitree G1.}
    \label{fig:side_flip_hardware}
\end{figure}

Second, we formulate a kinodynamic TO problem to recover dynamically feasible motion references from the calibrated kinematic sequence. Rather than solving this highly nonlinear and nonconvex kinodynamic TO problem~\cite{khazoom2024tailoring} in one shot, we tackle it progressively in three stages: kinematic TO, inverse dynamics, and full kinodynamic TO. Each stage uses the optimal solution of the previous stage as an initial guess, gradually introducing kinematic and dynamic consistency. The resulting trajectories provide not only dynamically valid state trajectories but also corresponding torque profiles, which serve as strong supervision for downstream RL-based tracking policies. This is especially beneficial when training highly dynamic motions like the side flip in \Fig{\ref{fig:side_flip_hardware}}.

Our approach bridges human demonstration, model-based TO, and learning-based control. By disentangling kinematic alignment, dynamic feasibility recovery, and robust control policy learning, we improve both motion quality and training efficiency. In summary, our main contributions are as follows.
\begin{itemize}
\item A general algorithm that generates and calibrates a human URDF from task-space motion, enabling physically interpretable human-to-robot geometric alignment and improved IK retargeting quality.
\item A progressive kinodynamic TO framework that gradually refines retargeted motions into kinematically and dynamically feasible humanoid trajectories.
\item An integrated pipeline that produces high-quality motion and torque references for RL training, bridging kinematic retargeting, dynamics enforcement through TO, and motion tracking controller training through RL.
\end{itemize}

\section{Related Work}
\subsection{Human Motion Retargeting}
\label{sec:related_work:human_motion_retarget}
Retargeting human motions to humanoid robots is essential for leveraging human motion datasets in robot learning and control. A central challenge arises from the morphological differences between humans and robots, including discrepancies in link lengths and kinematic structures.

One line of work addresses this issue through simplified task space modifications that are incompatible with the underlying skeleton structure. GMR~\cite{araujo2025retargeting}, for example, applies root-to-keyframe scalings and per-keyframe local offsets to human motion before using it as IK targets. While effective within a limited range of configurations, the correction is inherently local: a taller and wider shoulder and a shorter arm should be corrected by lengthening spine and shoulder links and shortening arm links. With root-to-arm scaling, however, the optimal scaling factor depends on joint configurations, making it hard to calibrate. GMR mitigates this issue by adding keyframe orientation tracking when the position references are off, which introduces nontrivial human-to-robot orientation offset calibration and additional IK tuning.

An alternative strategy is to calibrate the underlying human skeletal model to better match the robot dimensions. PHC~\cite{luo2023perpetual} optimizes the shape parameters of the human model SMPL to align it with the dimensions of the robot, thus producing task-space references that are more consistent with the target robot. However, the SMPL~\cite{loper2023smpl} shape parameters are derived from principal component analysis (PCA) on the shapes of human bodies. Although they are well-suited for representing realistic human body variations, calibrating them to robot dimensions, which are often outside the human distribution, can introduce undesirable artifacts, such as asymmetric limb proportions, spine distortions, or inconsistent foot geometry.

In contrast, our method directly calibrates the human URDF generated from task space human motion, which are not only more physically interpretable and accurate but also more generalizable to diverse human motion formats with an implicit skeleton structure other than SMPL.

\subsection{Model-Based Trajectory Optimization for Humanoids}
Model-based trajectory optimization (TO) serves as a cornerstone for generating dynamically feasible motions in high-dimensional humanoid systems. These methods formulate motion generation as a constrained optimization problem, enforcing physically grounded dynamics constraints ranging from simplified centroidal dynamics~\cite{dai2014whole} to full order Lagrangain dynamics~\cite{khazoom2024tailoring}. With careful engineering, such optimization problems can be used to control humanoid robots in real time using solving methods such as Differential Dynamic Programming~\cite{mayne1966second, OCS2} and Sequential Quadratic Programming~\cite{khazoom2024tailoring, nocedal2006numerical}.

Despite their success, classical model-based TO pipelines typically rely on carefully hand-crafted reference trajectories. These references include but are not limited to predefined base link pose and velocity trajectories~\cite{khazoom2024tailoring} and foot swing trajectories and contact schedules~\cite{OCS2}. Designing such references is non-trivial in that it requires substantial domain expertise, manual tuning for each robot morphology, and often task-specific adjustments. Moreover, because these references are hard coded rather than optimized, the resulting motions may appear overly regularized or conservative, lacking the variability, expressiveness, and subtle coordination patterns observed in natural human movements. As a result, even when dynamically feasible, the generated behaviors can be perceptually unnatural.

In contrast, our work starts from natural human motion references and refines them through model-based kinodynamic TO. We treat human demonstrations as rich priors and use TO to enforce robot-specific dynamics, contact feasibility, and actuation limits. This shifts the role of trajectory optimization from motion design to motion refinement, enabling dynamically consistent humanoid behaviors that retain the structure and naturalness of human movement while satisfying the physical constraints.

\subsection{Reinforcement Learning for Robust Motion Tracking}
Reinforcement learning (RL) has become a dominant paradigm for humanoid control due to its ability to learn feedback policies that are robust to modeling errors, disturbances, and contact uncertainties. However, RL-based control faces two persistent challenges: policies trained purely from task rewards may converge to unnatural motions, and training high-dimensional humanoid policies typically requires large amounts of data and computation~\cite{mittal2025isaac}.

To improve training efficiency and motion quality, prior works commonly adopt one of the following two strategies. A first approach uses model-based TO to synthesize dynamically feasible reference trajectories, and subsequently trains RL policies to track them~\cite{wu2023learning, liu2025opt2skill}. Although this reduces exploration difficulty and accelerates convergence, the reference motions remain shaped by handcrafted objectives, contact schedules, and heuristic design choices, inheriting the same naturalness limitations as pure TO methods. A second approach directly feeds kinematically retargeted human motions into RL as tracking references~\cite{peng2018deepmimic, peng2021amp, liao2025beyondmimic}. This significantly enhances motion naturalness by leveraging human demonstrations, but because kinematic retargeting does not enforce dynamics constraints, the RL policy must implicitly correct dynamics infeasibility. Consequently, motion correction and policy learning become entangled, increasing the difficulty of training.

In contrast, our approach bridges these paradigms. We first retarget natural human motion to the robot kinematics and then refine it through kinodynamic TO to enforce dynamics consistency. The optimized motions are subsequently used as references for RL training. This preserves the structural naturalness of human demonstrations while ensuring physical plausibility prior to policy learning. Moreover, our kinodynamic TO provides joint torque references that are unavailable in purely kinematic retargeting pipelines, offering richer supervision and further reducing learning complexity.

\section{Inverse Kinematics with URDF Calibration}
\begin{figure}
    \centering
    \includegraphics[width=1\linewidth]{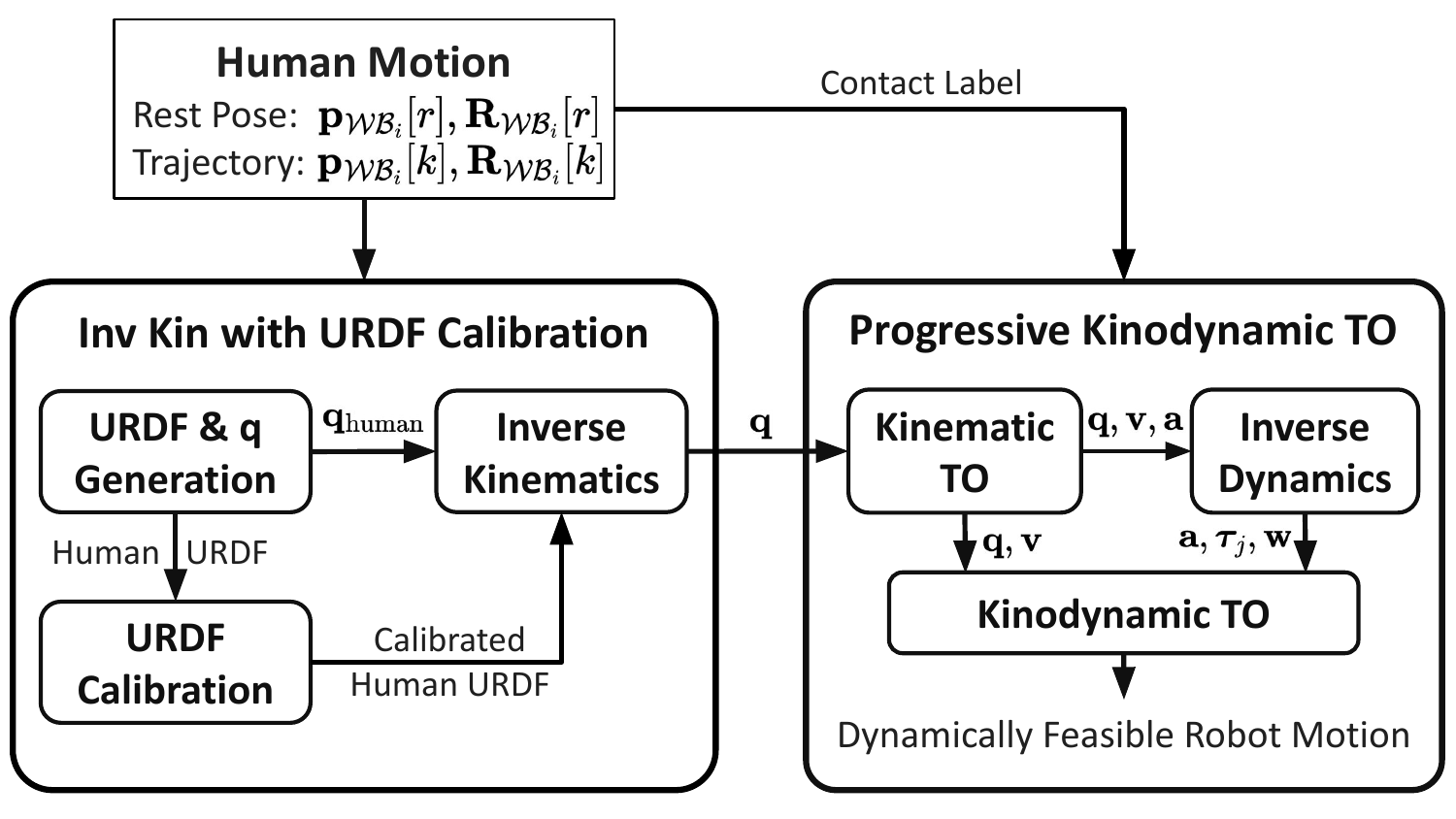}
    \caption{Retargeting framework.}
    \label{fig:framework}
\end{figure}

\Fig{\ref{fig:framework}} provides an overview of our Skeleton-Parameter Aligned Retargeting framework with Kinodynamic trajectory optimization (SPARK), which consists of an upstream inverse-kinematics (IK) stage and a downstream trajectory-optimization (TO) stage. In the IK stage, we convert task space human motion into a human URDF and a human generalized coordinate trajectory (\Sec{\ref{sec:urdf_q_gen}}). We then calibrate the human URDF to the target robot dimensions (\Sec{\ref{sec:urdf_calib}}), and replay the generalized coordinates on the calibrated URDF to generate task space references for IK (\Sec{\ref{sec:ik}}), producing a robot motion reference in robot generalized coordinates. In the TO stage, we refine this trajectory by removing kinematic artifacts via kinematic TO (KTO; \Sec{\ref{sec:kin_to}}) and dynamic artifacts via inverse dynamics (ID; \Sec{\ref{sec:inv_dyn}}) and kinodynamic TO (KDTO; \Sec{\ref{sec:kdto}}).

\subsection{Human URDF and Generalized Coordinates Generation}
\label{sec:urdf_q_gen}
\block{Human URDF Generation}
Human skeletons and humanoid robots typically differ in both the number of links and their dimensions. To enable reliable retargeting, we first convert a human skeleton motion sequence into a human URDF and a generalized-coordinate trajectory, and then calibrate the human URDF to match the target robot dimensions. Replaying the generalized coordinates on the calibrated URDF yields high-quality task-space references (\Sec{\ref{sec:ik}}) that are consistent with the robot morphology.
\begin{figure}
    \centering
    \includegraphics[width=\linewidth]{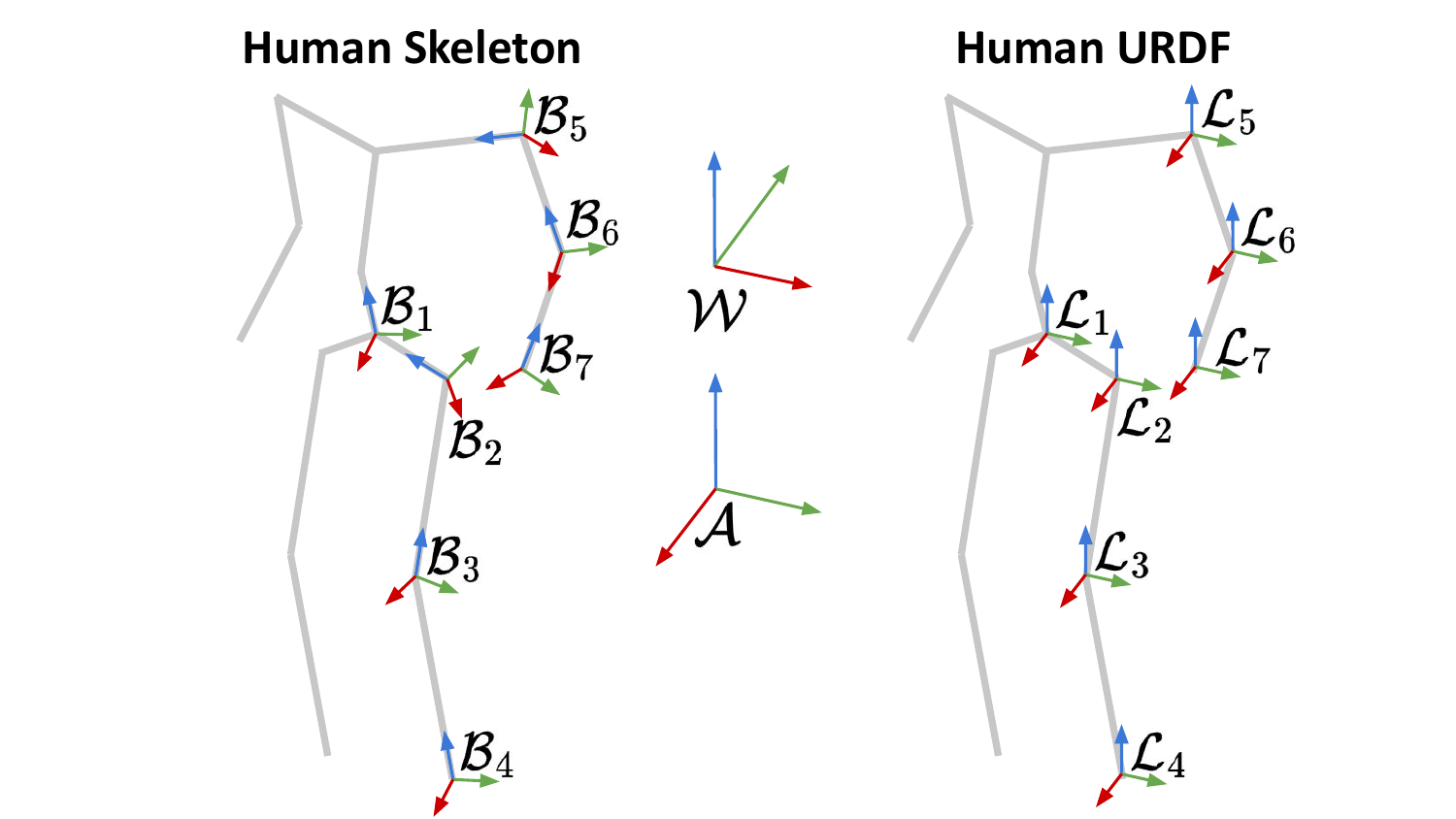}
    \caption{Coordinate frame definitions of human skeleton and the corresponding human URDF in rest pose. While the human bone frames $\cB_i$ are often aligned to the specific bones, we align the URDF link frames $\cL_i$ to a user-defined aligned frame $\cA$ with the same orientation as the robot URDF root link to facilitate URDF calibration.}
    \label{fig:human_skeleton_to_urdf}
\end{figure}

The conversion workflow is illustrated in \Fig{\ref{fig:human_skeleton_to_urdf}}. A human skeleton motion sequence provides the pose of each bone frame $\cB_i$ $(i\in\{1,\ldots,N_\txtframe\})$ with respect to the world frame $\cW$. In the URDF, we create one link $\cL_i$ per bone $\cB_i$ and construct the model from a rest pose specified by the motion file. The rest pose is typically a standing configuration with vertical legs and an upright torso; arm configurations may vary but should remain left--right symmetric. If a suitable rest pose is missing, we define one prior to URDF construction.

The coordinate-frame-related notations in this paper are as follows: For any three frames $\cA, \cB, \cC$, ${}^{\cC}\bp_{\cA\cB}$ denotes the position vector from the origin of frame $\cA$ to the origin of frame $\cB$, expressed in frame $\cC$. $\bR_{\cA\cB}$ denotes the rotation matrix that takes in a vector $\bv$ expressed in $\cB$ and returns the same vector expressed in $\cA$, i.e. ${}^{\cA}\bv = \bR_{\cA\cB} {}^{\cB}\bv$. It is also the relative rotation of frame $\cB$ with respect to frame $\cA$. $\bxi_{\cA\cB}$ denotes the quaternion corresponding to $\bR_{\cA\cB}$.

In the rest pose $r$, we set each link origin to match the corresponding bone origin, i.e., $\bp_{\cW\cL_i}[r]=\bp_{\cW\cB_i}[r]$. We then rotate each link frame to coincide with the orientation of a user-defined aligned frame $\cA$, which has the same orientation as the robot URDF root link to facilitate the URDF calibration in \Sec{\ref{sec:urdf_calib}}. This alignment simplifies both URDF generation and calibration: the local rest-pose relative translations ${}^{\cL_i}\bp_{\cL_i\cL_j}[r]$ become identical to the global relative translations expressed in $\cA$, i.e., ${}^{\cL_i}\bp_{\cL_i\cL_j}[r] = {}^{\cA}\bp_{\cL_i\cL_j}[r]$. Consequently, constructing the URDF reduces to computing these relative translations from the world-frame rest-pose bone positions ${}^{\cW}\bp_{\cW\cB_i}[r]$ using
\begin{equation}
    \begin{aligned}
        {}^{\cL_{i}}\bp_{\cL_{i}\cL_{j}}[r]
        &= {}^\cA\bp_{\cL_{i}\cL_{j}}[r] \\
        &= \bR_{\cW\cA}^T\left(
            {}^\cW\bp_{\cW\cL_{j}}[r] - {}^{\cW}\bp_{\cW\cL_{i}}[r]
        \right) \\
        &= \bR_{\cW\cA}^T\left(
            {}^\cW\bp_{\cW\cB_{j}}[r] - {}^{\cW}\bp_{\cW\cB_{i}}[r]
        \right).
    \end{aligned}
    \label{eq:urdf_gen}
\end{equation}

\block{Human Generalized Coordinate Generation}
We define the generalized coordinates of the human URDF at timestep $k$ as
$\bq_{\human}[k]=\left[{}^\cW\bp_{\cW\cL_{1}}[k]; \bxi_{\cW\cL_{1}}[k]; \bphi_1[k]; \ldots; \bphi_{N_j}[k]\right]$, where ${}^\cW\bp_{\cW\cL_{1}}$ is the root-link position, $\bxi_{\cW\cL_{1}}$ is the root-link quaternion, and $\bphi_n$ $(n\in\{1,\ldots,N_j\})$ are the Euler angles of the $n$-th spherical joint. These quantities are computed as
\begin{subequations}
    \begin{align}
        {}^\cW\bp_{\cW\cL_{1}}[k]
        &= s_\txtroot{}^\cW\bp_{\cW\cB_{1}}[k], \label{eq:q_traj_gen:root_pos}\\
        \bxi_{\cW\cL_{1}}[k]
        &= \quat\left(
            \bR_{\cW\cL_{1}}[k]
        \right) \nonumber \\
        &= \quat\left(
            \bR_{\cW\cB_{1}}[k]\bR_{\cB_{1}\cL_{1}}
        \right), \\
        \bphi_n[k] &= \euler\left(
            \bR_{\cL_{i_n}\cL_{j_n}}[k]
        \right) \nonumber \\
        &= \euler\left( \bR_{\cB_{i_n}\cL_{i_n}}^T\bR_{\cB_{i_n}\cB_{j_n}}[k]\bR_{\cB_{j_n}\cL_{j_n}}
        \right), \end{align}\label{eq:q_traj_gen}\end{subequations}
where $s_\txtroot$ is the root position scaling factor to be calibrated in \Sec{\ref{sec:urdf_calib}}, $\quat(\cdot)$ converts rotation matrix to quaternion, and $\euler(\cdot)$ converts rotation matrix to euler angles. $\bR_{\cB_{i}\cL_{i}}$ is the relative rotation of the $i$-th link frame with respect to the corresponding bone frame, which can be computed under rest pose as
\begin{equation}
    \bR_{\cB_{i}\cL_{i}} = \bR_{\cB_{i}\cA}[r] = \bR_{\cW\cB_i}^T[r]\bR_{\cW\cA}.
    \label{eq:bone_to_link_rot}
\end{equation}

\subsection{Human URDF Calibration}
\label{sec:urdf_calib}
We calibrate the human URDF to the target humanoid dimensions by matching keyframes in the respective rest poses. \Fig{\ref{fig:urdf_calib}} shows an example of the rest pose configurations and key frame definitions when calibrating human URDF to Unitree G1. The robot rest pose (typically the neutral configuration with zero joint angles) should be a standing pose compatible with the human-URDF rest pose convention. The arm configuration may differ, since we only calibrate arm lengths. We decompose calibration into five groups:
\begin{figure}
    \centering
    \includegraphics[width=\linewidth]{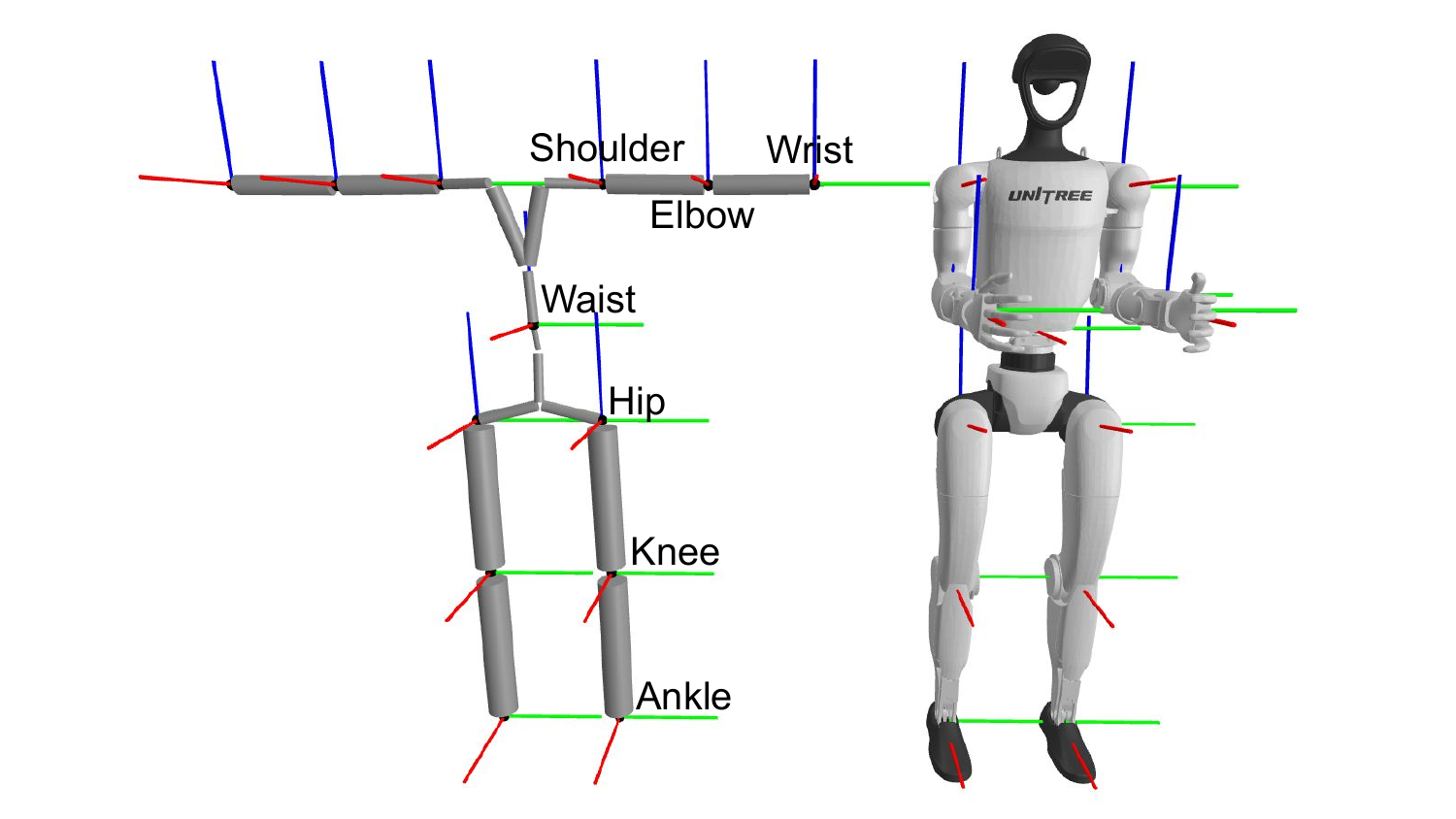}
    \caption{Rest pose key frame calibration of human URDF to humanoid robot dimensions. We match the link lengths for arms regardless of their absolute positions, and the absolute positions for the remaining key frames. We also calibrate the orientation offsets for end effectors if the robot has enough degree of freedom to track end effector orientations.}
    \label{fig:urdf_calib}
\end{figure}

\block{End Effector Calibration}
We calibrate the local position and orientation offsets of end effectors such as wrists and ankles from the human URDF to the robot URDF so that their absolute poses can be tracked. If the robot wrist has fewer than three rotational degrees of freedom, its orientation cannot be tracked exactly; in this case, we omit the wrist-orientation calibration. In our IK formulation, end effectors are the only frames with orientation targets, since orientation offsets for intermediate links are generally ambiguous.

\block{Arm Calibration}
We select three keyframes on each arm (shoulder, elbow, and wrist) and calibrate the shoulder--elbow and elbow--wrist link lengths. Because human and robot rest poses can differ substantially in arm configuration, we do not calibrate the full 3D relative translations. Instead, we estimate a scalar length scaling factor $s_\length$ for each segment through the following
\begin{equation}
    s_\length = \LtwoNorm{\bp_\robot} / \LtwoNorm{\bp_{\human}},
    \label{eq:urdf_calib:link_length}
\end{equation}
where $\LtwoNorm{\bp_\robot}$ is the arm length of the robot and $\LtwoNorm{\bp_\human}$ is the arm length of the human.

\block{Leg Calibration}
We choose three keyframes on each leg (hip, knee, and ankle) and calibrate the absolute hip--knee and knee--ankle translations in the human URDF to match the robot URDF.

\block{Main Body Calibration}
We select five keyframes on the main body (one waist frame, two shoulder frames, and two hip frames). Because the torso contains additional intermediate links, we correct their dimensions using an affine transformation. In the Unitree G1 example shown in \Fig{\ref{fig:urdf_calib}}, where the robot URDF root frame and all human URDF frames have their x axes pointing forward and z axes pointing upward in the rest pose, this affine transformation includes an $xz$ shear and independent scalings along $y$ and $z$. The orientation synchronization was done by the aligned frame $\cA$ introduced in \Sec{\ref{sec:urdf_q_gen}}. To calibrate the rest pose transformations of all links between $\cL_i$ and $\cL_j$ in the human URDF, we find their corresponding links $\cR_i, \cR_j$ in the robot URDF, and calibrate the affine transformation $\bA$ using
\begin{equation}
     \bp_{\cR_i\cR_j}
     = \bA \bp_{\cL_i\cL_j}
     = \left[\begin{matrix}
         1 & 0 & k_x \\
         0 & s_y & 0 \\
         0 & 0 & s_z
     \end{matrix}\right] \bp_{\cL_i\cL_j},
    \label{eq:urdf_calib:affine_transf}
\end{equation}
where $k_x$ is the xz shear factor and $s_y, s_z$ are the scaling factors along the y and z directions. We then apply the calibrated transform to any intermediate link relative translation $\bp_{\cL_m\cL_n}$, where $\cL_m$ and $\cL_n$ are links between $\cL_i$ and $\cL_j$, using
\begin{equation}
    \bp_{\cL_m\cL_n,\corrected} = \bA\bp_{\cL_m\cL_n}.
\end{equation}

Setting $\cL_i$ to the waist frame and $\cL_j$ to a shoulder frame calibrates all links between waist and shoulders. Likewise, setting $\cL_i$ to the waist frame and $\cL_j$ to a hip frame calibrates links between waist and hips.

\block{Root Position Scale Calibration}
After calibrating leg and main body dimensions, we need to calibrate the root position scaling factor $s_\txtroot$ in \eqref{eq:q_traj_gen:root_pos} so that stance legs are still sticking on ground after leg dimensions have changed. The calibration can be done using
\begin{equation}
    s_\txtroot = \LtwoNorm{\bp_{\cL_\txtroot\cL_\ankle}^\corrected} / \LtwoNorm{\bp_{\cL_\txtroot\cL_\ankle}},
\end{equation}
where $\bp_{\cL_\txtroot\cL_\ankle}^\corrected$ and $\bp_{\cL_\txtroot\cL_\ankle}$ are the corrected and original root-to-ankle vector respectively.

\subsection{Inverse Kinematics (IK)}
\label{sec:ik}
After URDF calibration (\Sec{\ref{sec:urdf_calib}}), we replay the human generalized-coordinate trajectory in \eqref{eq:q_traj_gen} on the calibrated human URDF to obtain task-space references for all keyframes. Because calibration aligns the reference geometry with the target robot morphology, these task-space references can be tracked reliably. We solve IK for the generalized coordiantes $\bq$ of the humanoid robot at each timestep via
\begin{equation}
    \begin{aligned}
        \min_{\bq} \;&
        \sum_{i=1}^{N_\EE}\WeightedLtwoNormSquare{\delta\bR(\bq)}{\bW_{\bR}} + \sum_{i=1}^{N_\txtframe}\WeightedLtwoNormSquare{\delta\bp(\bq)}{\bW_{\bp}} \\ &+
        \WeightedLtwoNormSquare{\bq_j - \bq_{j,\prev}}{\bW_{\bq_j}} \\
        \mathrm{s.t.} \;
        & \bq_{j,\min} \leq \bq_j \leq \bq_{j,\max} \\
        & \bv_{j,\min}\Delta t \leq \bq_j - \bq_{j,\prev} \leq \bv_{j,\max} \Delta t,
    \end{aligned}
    \label{eq:ik_formulation}
\end{equation}
where $\WeightedLtwoNormSquare{\be}{\bW}:=\be^T\bW\be$ denotes a $\bW$-weighted squared $\ell_2$ norm. The objective minimizes position errors $\delta\bp$ over all $N_\txtframe$ keyframes and orientation errors $\delta\bR$ only over the $N_\EE$ end effectors. We omit orientation targets for intermediate links, since their orientations are determined by the connected keyframe positions. This avoids calibrating ambiguous intermediate-link orientation offsets. If a robot wrist has fewer than three rotational degrees of freedom, we also drop wrist-orientation tracking. Finally, we enforce joint position and velocity limits and regularize towards the previous solution $\bq_{\prev}$ to reduce jitter near singularities.

\section{Progressive Kinodynamic TO}
\subsection{General TO Formulation}
Given the IK retargeting result $\bq_\IK^*$ and the contact labels from the human motion, we further apply kinodynamic trajectory optimization (KDTO) to obtain humanoid motions that are feasible both kinematically and dynamically. Directly solving KDTO is difficult due to the problem's high dimensionality and nonlinearity, so we adopt a progressive pipeline with three stages: (i) kinematic TO (KTO; \Sec{\ref{sec:kin_to}}) to correct contact kinematics and self-collisions, (ii) per-timestep inverse dynamics (ID; \Sec{\ref{sec:inv_dyn}}) to produce a good initialization for second-order dynamic quantities, and (iii) a final KDTO stage (\Sec{\ref{sec:kdto}}) that jointly optimizes kinematics and dynamics. Both KTO and KDTO can be written in the following general form:
\begin{subequations}
    \begin{align}
        \min_{\bx[0:N], \bu[0:N-1]} \;&
        l_\track(\bx[0:N]) + l_\reg(\bx[0:N], \bu[0:N-1]) \nonumber \\
        \mathrm{s.t.} \;&
        \bx[k+1] = \boldf\left(\bx[k], \bu[k]\right)
        \label{eq:gen_to_formulation:sys_dyn} \\
        & \bg(\bx[k]) \leq \bzero \nonumber \\
        & \bh\left(\bx[k], \bu[k] \right) \leq \bzero \label{eq:gen_to_formulation:ineq_constr} \\
        & \bx_{\min} \leq \bx[k] \leq \bx_{\max} \nonumber \\
        & \bu_{\min} \leq \bu[k] \leq \bu_{\max}.
        \label{eq:gen_to_formulation:box_constr}
\end{align}\label{eq:gen_to_formulation}\end{subequations}
The exact definitions of the decision variables, costs and constraints differ between KTO (\Sec{\ref{sec:kin_to}}) and KDTO (\Sec{\ref{sec:kdto}}), and are defined in the corresponding subsections. In general, the decision variables are the system states $\bx$ and inputs $\bu$. The objective consists of a motion-tracking term $l_\track(\bx[0:N])$ and a smoothing regularizer $l_\reg(\bx[0:N],\bu[0:N-1])$. The constraints include the system dynamics \eqref{eq:gen_to_formulation:sys_dyn}, general inequality constraints \eqref{eq:gen_to_formulation:ineq_constr}, and box constraints \eqref{eq:gen_to_formulation:box_constr} on states and inputs. Constraints involving time index $k+1$ apply for all $k\in\{0,\ldots,N-1\}$; all others apply for $k\in\{0,\ldots,N\}$.

\subsection{Kinematic TO (KTO)}
\label{sec:kin_to}
\block{Decision Variables} For KTO, the system state is $\bx=[\bq;\bv]$, comprising the robot generalized coordinates $\bq$ and generalized velocities $\bv$, and the system input is the generalized acceleration $\bu=\ba$. We initialize KTO using the IK solution: $\bq_{\guess}=\bq_\IK^*$. We then obtain $\bv_{\guess}$ and $\ba_{\guess}$ via finite differences.

\block{Cost Function}
The KTO tracking cost $l_{\track}^{\KTO}$ penalizes task-space pose errors $\delta\bp_i,\delta\bR_i$ and task-space velocity errors $\delta\bv_i,\delta\bomg_i$ for all $N_\txtframe$ keyframes used in IK, with targets computed from the IK solution $\bq_\IK^*$ and $\bv_\IK^*$. The regularization term $l_{\reg}^{\KTO}$ penalizes generalized accelerations $\ba$ to smooth the trajectory.
\begin{subequations}
    \begin{align}
        l_{\track}^{\KTO} &=
        \sum_{k=0}^N\sum_{i=0}^{N_{\txtframe}}
        \WeightedLtwoNormSquare{\delta\bp_{i}(\bq[k])}{\bW_\bp} +
        \WeightedLtwoNormSquare{\delta\bR_{i}(\bq[k])}{\bW_\bR} + \nonumber \\
        & \WeightedLtwoNormSquare{\delta\bv_{i}(\bq[k], \bv[k])}{\bW_\bv} +
        \WeightedLtwoNormSquare{\delta\bomg_{i}(\bq[k], \bv[k])}{\bW_{\bomg}}
        \label{eq:kto_cost:track} \\
        l_\reg^\KTO &= \sum_{k=0}^{N-1}\WeightedLtwoNormSquare{\ba[k]}{\bW_\ba}
        \label{eq:kto_cost:reg}
    \end{align}
    \label{eq:kto_cost}
\end{subequations}

\block{Double-Integrator Dynamics Constraint} The KTO dynamics use a semi-implicit double integrator that maps generalized accelerations $\ba$ to states $(\bq,\bv)$:
\begin{equation}
    \begin{aligned}
        \bq[k+1] &= \bq[k] \oplus \bv[k+1] \,\Delta t \\
        \bv[k+1] &= \bv[k] + \ba[k] \,\Delta t,
    \end{aligned}
    \label{eq:kto_constr:double_integrator}
\end{equation}
where $\oplus$ denotes the additive group operation (to handle configuration manifolds).

\block{Kinematic Contact Constraints}
A key role of KTO is to enforce consistent contact kinematics. We impose three sets of constraints:
\begin{subequations}
    \begin{align}
        p_{\swing, z} \geq 0,
        \label{eq:kto_constr:contact_kin:swing_ft_above_gnd} \\
        p_{\text{first contact}, z} = 0 \nonumber \\
        R_{\text{first contact}, zz} = 1 - \epsilon ,
        \label{eq:kto_constr:contact_kin:first_contact_on_gnd} \\
        \bv_\contact = \bzero \nonumber \\
        \bomg_\contact = \bzero.
        \label{eq:kto_constr:contact_kin:holonomic_contact}
    \end{align}
    \label{eq:kto_constr:contact_kin}
\end{subequations}
The swing-foot-above-ground constraint \eqref{eq:kto_constr:contact_kin:swing_ft_above_gnd} enforces a nonnegative swing-foot height $p_{\swing,z}$. The first-contact-on-ground constraint \eqref{eq:kto_constr:contact_kin:first_contact_on_gnd} sets the contact-foot height to zero at the first contact timestep and approximately aligns the foot with the ground by constraining $R_{\text{first contact},zz}=1-\epsilon$, where $\epsilon$ is a small slack to avoid numerical infeasibility (we use $\epsilon=10^{-4}$). Finally, the holonomic contact constraint \eqref{eq:kto_constr:contact_kin:holonomic_contact} enforces zero linear and angular velocities $\bv_\contact, \bomg_\contact$ at the contact foot. Together, these constraints keep the swing foot above the ground and establish holonomic surface contact for the stance foot.

\block{Self Collision Avoidance Constraint}
We formulate self-collision avoidance constraints on a point-to-point basis. At any timestep $k$, for each collision pair $(\bp_i,\bp_j)\in\cC$ (with $i,j$ denoting collision-point indices), we enforce that the distance between $\bp_i$ and $\bp_j$ exceeds a minimum separation $d_{(i,j),\min}$:
\begin{equation}
    \LtwoNorm{\bp_{i}(\bq[k]) - \bp_{j}(\bq[k])} \geq d_{(i,j),\min}.
    \label{eq:kto_constr:self_collision}
\end{equation}
This formulation extends naturally to sphere-to-sphere constraints by increasing $d_{(i,j),\min}$ by the sum of radii.

\block{Joint Position and Velocity Limits}
The KTO box constraints include joint-position limits and joint-velocity limits:
\begin{equation}
    \begin{aligned}
        \bq_{j, \min} &\leq \bq_{j}[k] \leq \bq_{j, \max}, \\
        \bv_{j, \min} &\leq \bv_{j}[k] \leq \bv_{j, \max}.
    \end{aligned}
    \label{eq:kto_constr:jpos_jvel_lim}
\end{equation}

\subsection{Inverse Dynamics (ID)}
\label{sec:inv_dyn}
At each timestep, we fix the generalized coordinates and velocities to the KTO optimum $(\bq_\KTO^*,\bv_\KTO^*)$ and solve inverse dynamics (ID) as a single-timestep quadratic program (QP).

\block{Decision Variable}
ID optimizes the generalized acceleration $\ba$, joint torques $\btau_j$, and contact wrenches $\bw$. The KTO-optimal acceleration $\ba_\KTO^*$ serves as both the tracking reference and the initialization for $\ba$.

\block{Cost Function}
The ID objective tracks the KTO-optimal acceleration $\ba_\KTO^*$ while minimizing joint torques $\btau_j$ and contact wrenches $\bw$:
\begin{equation}
    \begin{aligned}
        l^\ID &= \WeightedLtwoNormSquare{\delta\ba}{\bW_\ba} + 
        \WeightedLtwoNormSquare{\btau_j}{\bW_{\btau_j}} +
        \WeightedLtwoNormSquare{\bw}{\bW_\bw}.
    \end{aligned}
    \label{eq:id_cost}
\end{equation}

\block{Lagrangian Dynamics Constraint}
The ID dynamics constraint uses the full-order Lagrangian rigid body dynamics of a floating-base system:
\begin{equation}
    \bM(\bq)\ba + \bH(\bq,\bv) =
    \left[\begin{matrix}
        \bzero \\
        \btau_{j}
    \end{matrix}\right] +
    \sum_{i=1}^{N_\text{foot}}\bJ_{i}^T(\bq)\bw_{i}.
    \label{eq:id_constr_rnea}
\end{equation}

\block{Contact Wrench Constraints}
When a foot is in swing phase, its contact wrench must vanish, i.e. $\bw_\swing = \bzero$. When a foot is in contact phase, the 6D wrench of a foot $\bw = [f_x; f_y; f_z; \tau_x; \tau_y; \tau_z]$ satisfies the Contact Wrench Cone (CWC) constraint~\cite{caron2015stability}, which can be written as:
\begin{equation}
    \begin{aligned}
        |f_x| \le \mu f_z, \; |f_y| &\le \mu f_z, \; f_z > 0 \\
        |\tau_x| \le Y f_z, &\; |\tau_y| \le X f_z \\
        \tau_{z,\min} \le \tau_z &\le \tau_{z,\max},
    \end{aligned}
    \label{eq:id_constr:cwc}
\end{equation}
\begin{equation}
    \begin{aligned}
        \tau_{z,\min} 
        &= 
        -\mu (X + Y) f_z 
        + \left| Y f_x - \mu \tau_x \right|
        + \left| X f_y - \mu \tau_y \right|, \\
        \tau_{z,\max} 
        &=
        +\mu (X + Y) f_z 
        - \left| Y f_x + \mu \tau_x \right|
        - \left| X f_y + \mu \tau_y \right|,
    \end{aligned}
    \label{eq:id_constr:cwc:yaw_torq_lim}
\end{equation}
where $X$ and $Y$ are the length and width of the foot support-box, and $\mu$ is the friction coefficient.

\block{Contact Foot Zero Acceleration Constraint}
In \eqref{eq:kto_constr:contact_kin}, the contact-foot velocity is constrained to zero, which implies zero contact-foot acceleration. However, because ID treats generalized acceleration $\ba$ as a decision variable and does not explicitly include \eqref{eq:kto_constr:contact_kin}, the resulting solution $\ba_\ID^*$ may violate this zero-acceleration condition. We therefore enforce it explicitly as:
\begin{equation}
    \bJ_\contact(\bq) \ba + \bJdot_\contact(\bq, \bv) \bv = \bzero.
\end{equation}

\block{Torque Limits}
The ID is subjected to joint torque limits:
\begin{equation}
    \begin{aligned}
        \btau_{j, \min} &\leq \btau_{j} \leq \btau_{j, \max}.
    \end{aligned}
    \label{eq:id_constr:tau_lim}
\end{equation}

\subsection{Kinodynamic TO (KDTO)}
\label{sec:kdto}
\block{Decision Variable}
The KDTO state is $\bx=[\bq;\bv;\ba]$, which includes generalized coordinates, velocities, and accelerations, and the input is $\bu=[\btau_j;\bw]$, which includes joint torques and contact wrenches. Since KDTO is highly nonlinear and nonconvex, we warm-start it using the KTO and ID solutions. Specifically, we initialize generalized coordinates and velocities with KTO: $\bq_{\guess}=\bq_\KTO^*$ and $\bv_{\guess}=\bv_\KTO^*$, and initialize accelerations, torques, and contact wrenches with ID: $\ba_{\guess}=\ba_\ID^*$, $\btau_{j,\guess}=\btau_{j,\ID}^*$, and $\bw_{\guess}=\bw_\ID^*$.

\block{Cost Function}
The KDTO costs share the same tracking term as KTO ($l_\track^\KDTO = l_\track^\KTO$), with a regularization term
\begin{equation}
    l_\reg^\KDTO =
    \sum_{k=0}^{N-1}
    \WeightedLtwoNormSquare{\delta\ba[k]}{\bW_\ba} +
    \WeightedLtwoNormSquare{\delta\btau_{j}[k]}{\bW_{\btau_j}} +
    \WeightedLtwoNormSquare{\delta\bw[k]}{\bW_\bw}.
\end{equation}
The regularization term encourages generalized accelerations $\ba$, joint torques $\btau_j$, and contact wrenches $\bw$ to stay close to the ID solutions $\ba_\ID^*$, $\btau_{j,\ID}^*$, and $\bw_\ID^*$.

\block{Constraints}
The KDTO constraints combine the KTO constraints with the ID constraints. For dynamics, we enforce both the discrete-time double-integrator constraint \eqref{eq:kto_constr:double_integrator} and the rigid-body dynamics constraint \eqref{eq:id_constr_rnea} at each timestep. For contact, we include the kinematic contact constraints \eqref{eq:kto_constr:contact_kin}, the zero-wrench constraint for swing foot, and the CWC constraint \eqref{eq:id_constr:cwc} for stance foot. Self-collision avoidance follows \eqref{eq:kto_constr:self_collision}. Finally, we impose joint position and velocity limits \eqref{eq:kto_constr:jpos_jvel_lim} and joint-torque limits \eqref{eq:id_constr:tau_lim}.

\section{Results and Analyses}
This section evaluates the key components of our pipeline through three sets of experiments. First, we quantify how URDF calibration improves IK retargeting accuracy on multiple humanoid platforms in \Sec{\ref{sec:effect_of_urdf_calib_on_ik}}. Next, we study motion editing in \Sec{\ref{sec:motion_edit}}, using forward jumping as a case study, and compare KDTO against raw edits and KTO. Finally, we demonstrate the advantage of KDTO for highly dynamic motions such as the side flip shown in \Fig{\ref{fig:side_flip_hardware}} in \Sec{\ref{sec:dyn_motion_track}}, where recovering dynamic feasibility and optionally providing torque references accelerates the RL training process.

\subsection{Effect of URDF Calibration on IK}
\label{sec:effect_of_urdf_calib_on_ik}
\begin{figure}
    \centering
    \includegraphics[width=\linewidth]{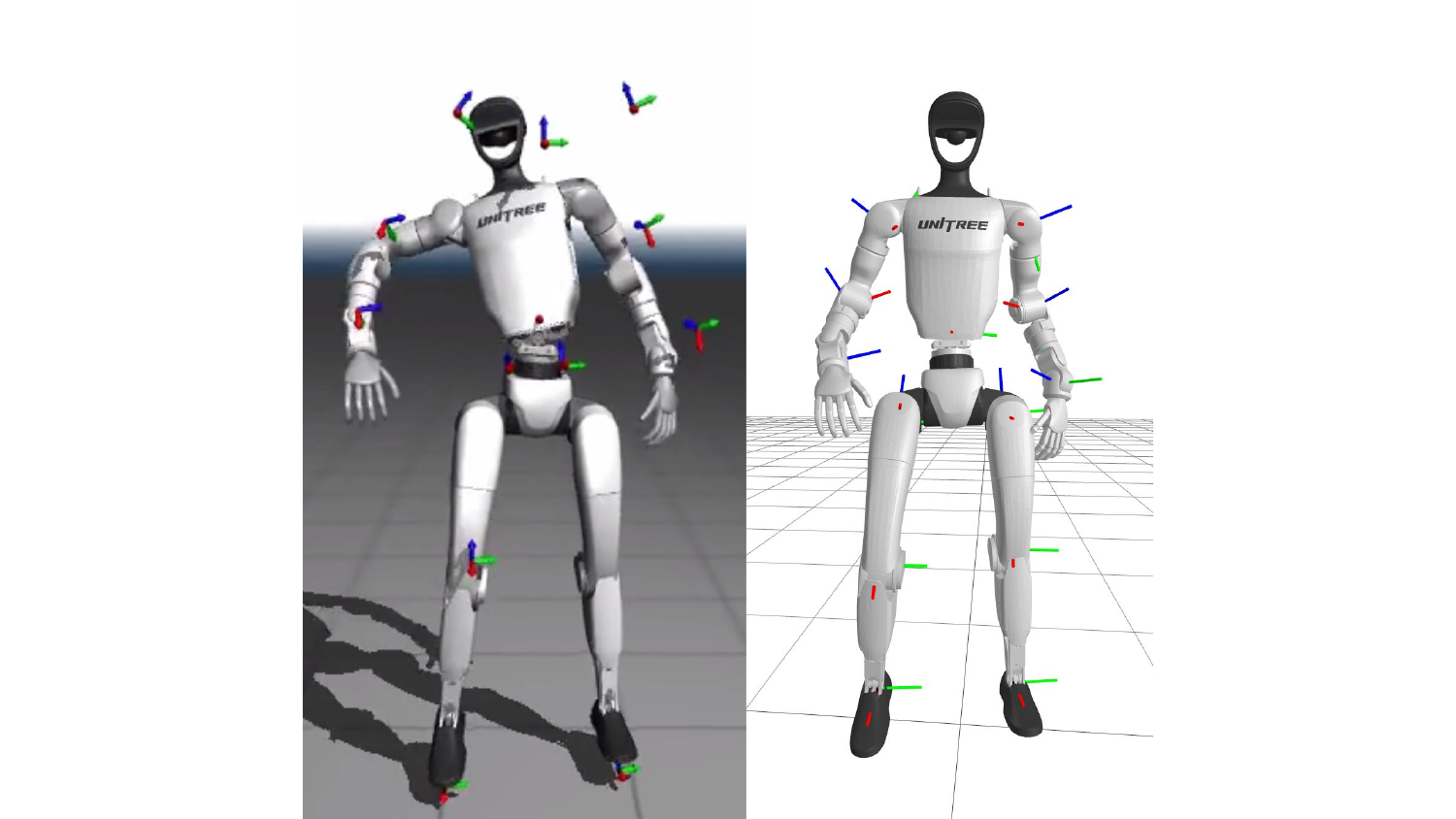}
    \caption{IK Comparison between GMR (left) and URDF Calibration (right).}
    \label{fig:gmr_vs_urdf_calib}
\end{figure}

As discussed in \Sec{\ref{sec:related_work:human_motion_retarget}}, prior retargeting methods such as GMR~\cite{araujo2025retargeting} model human--robot morphology differences using task-space root-to-keyframe scalings and per-keyframe local offsets that are not consistent with the underlying skeletal structure. Calibrating these scaling and offset parameters is therefore intertwined with tuning the IK cost, which makes the approach difficult to generalize across large variations in source human body shape and motion. When applied to datasets with substantial diversity in human shapes and motions, such as AMASS~\cite{AMASS:ICCV:2019}, it can fail in corner cases such as \Fig{\ref{fig:gmr_vs_urdf_calib}}, where the reference matches along one axis (width) but is incorrect along another (height).

Our calibration of URDF resolves this issue by correcting the dimensions of the actual human skeleton. Once the target robot and the human-motion format are fixed, the same calibration logic can be applied to large datasets containing human skeletons of various dimensions. Because the calibrated task space targets are structurally consistent with the robot, we do not need to impose orientation tracking on intermediate links, and the IK position and orientation tracking weights can be set uniformly to one. This substantially reduces the IK tuning effort. We benchmark the IK mean per-body position error $E_\mpbpe$ on the AMASS ACCAD dataset~\cite{AMASS:ICCV:2019} when retargeting to Unitree G1~\cite{UnitreeG1}, H1~\cite{UnitreeH1}, Booster T1~\cite{BoosterT1}, EngineAI PM01~\cite{EngineAIPM01}, and Kuavo 4Pro~\cite{Kuavo4Pro}. \Table{\ref{tab:ik_err}} shows that URDF calibration yields significantly lower $E_\mpbpe$ than GMR, with improvements of $82.9\%$ on G1, $64.9\%$ on H1, $71.9\%$ on T1, $74.0\%$ on PM01, and $75.8\%$ on Kuavo 4Pro.
\begin{table}
\centering
\caption{IK $E_\mpbpe$ [cm] Comparison between GMR and Our URDF Calibration on AMASS ACCAD Dataset over Multiple Robots.}
\begin{tabular}{@{}llllll@{}}
\toprule
Retarget Method  & G1    & H1    & T1    & PM01 & Kuavo 4Pro\\ \midrule
GMR              & 9.37          & 12.53         & 6.59          & 9.51 & 19.50\\
URDF Calibration & \textbf{1.60} & \textbf{4.40} & \textbf{1.85} & \textbf{2.47} & \textbf{4.71}\\ \bottomrule
\end{tabular}
\label{tab:ik_err}
\end{table}

\subsection{Motion Editing}
\label{sec:motion_edit}
\begin{figure}
    \centering
    \includegraphics[width=\linewidth]{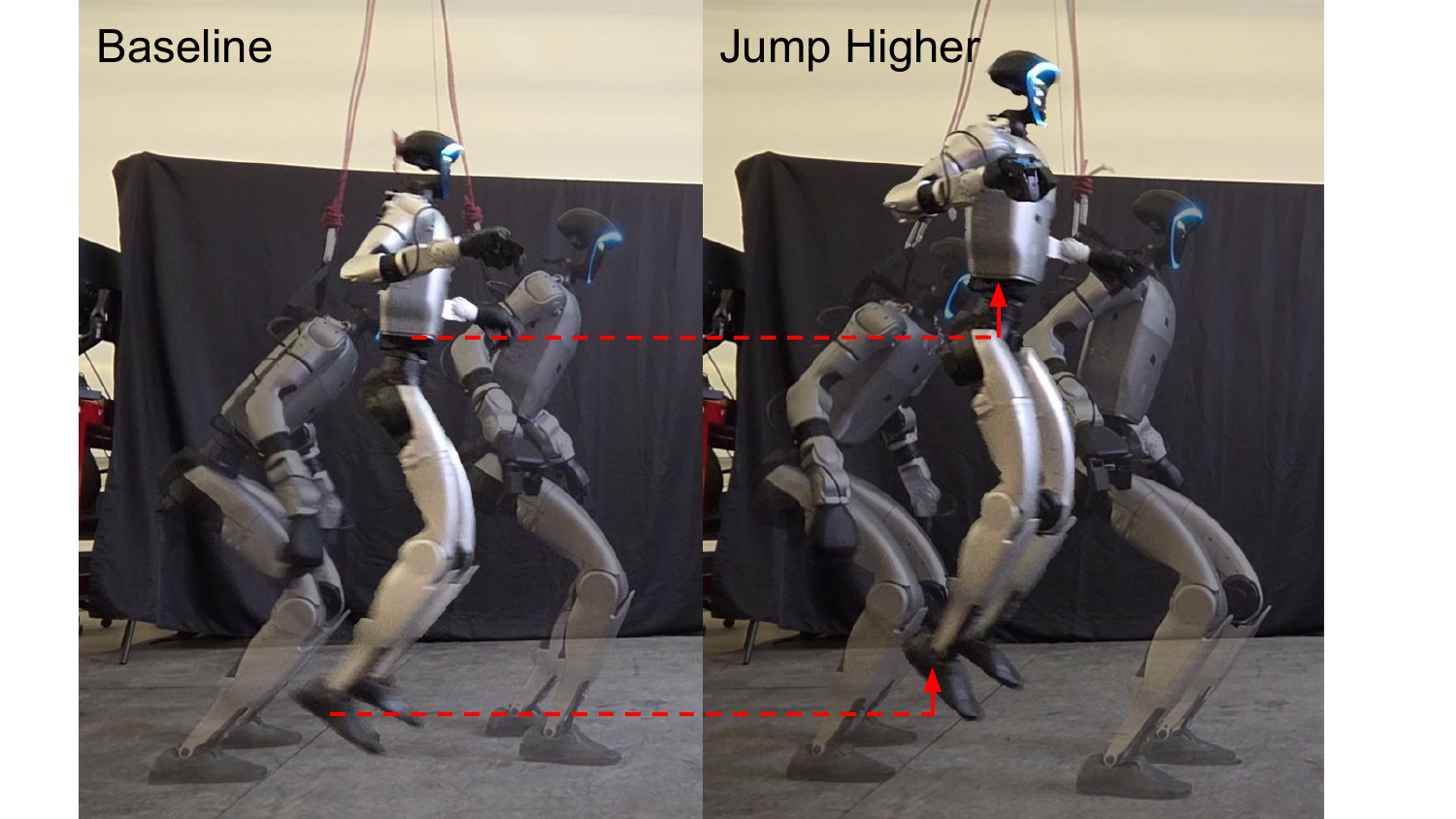}
    \caption{Unitree G1 performing a forward jump. The left case is the baseline, and the right one is edited to jump higher.}
    \label{fig:jump_higher_hardware}
\end{figure}

\begin{figure}
    \centering
    \includegraphics[width=\linewidth]{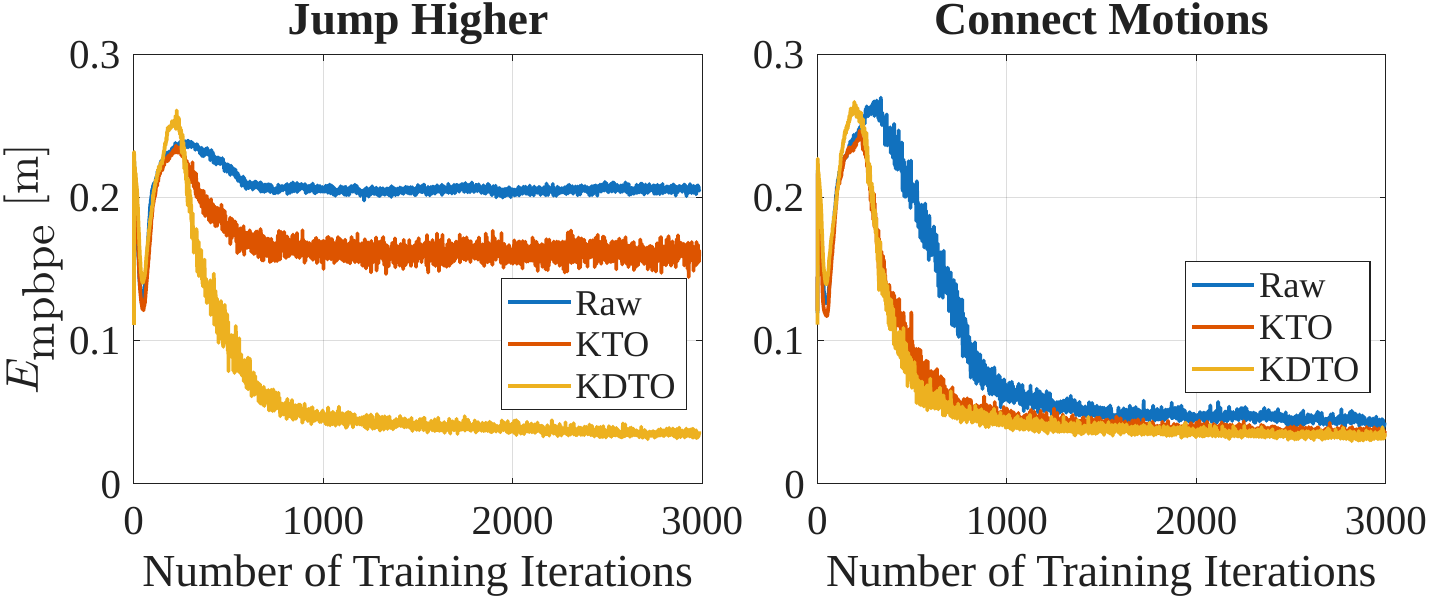}
    \caption{Comparison of RL $E_\mpbpe$ learning curves of jumping motion editing when the reference motions are provided by raw edit, KTO, and KDTO.}
    \label{fig:edit_raw_kto_kdto}
\end{figure}

Collecting human demonstrations is costly, so editing existing motions is an appealing way to synthesize additional training data and broaden hardware coverage. However, naive edits can introduce kinematic discontinuities and dynamic infeasibility, making the resulting references difficult to track. TO can mitigate these artifacts and improve reference quality. Here we use forward jumping as a case study. We apply two kinds of edits to it: (i) increasing the jump height shown in \Fig{\ref{fig:jump_higher_hardware}}, (ii) connecting it with an initial and terminal standing phase for safe hardware deployment. \Fig{\ref{fig:edit_raw_kto_kdto}} reports RL learning curves of the mean per-body position tracking error $E_\mpbpe$ when using references from the raw edit, KTO, and KDTO. We train motion-tracking policies following BeyondMimic~\cite{liao2025beyondmimic} in IsaacLab~\cite{mittal2025isaac} on Unitree G1~\cite{UnitreeG1}.

\block{Jump Higher}
Simply scaling the vertical position to obtain a higher jump is insufficient, because a larger apex height also requires a longer flight time under the same gravity. We therefore scale the foot $z$-height by $4\times$ and adjust the time discretization $\Delta t$ based on the resulting change in center-of-mass height. We measured that this height edit increases the maximum center-of-mass height by $1.4\times$, so we scale the time by $\sqrt{1.4}$. Instead of directly scaling $\Delta t$, which degrades the accuracy of TO integration, we interpolate the reference to $\sqrt{1.4}$ times the original frequency while keeping $\Delta t$ unchanged. The right subfigure of \Fig{\ref{fig:edit_raw_kto_kdto}} shows that KDTO achieves a substantially lower final tracking error than KTO and the raw edit. This highlights that when edits affect dynamics-relevant quantities such as gravitational acceleration, recovering a dynamically feasible trajectory with KDTO is important for high-quality tracking.

\block{Connect Motions}
The jumping motions cannot be deployed directly on hardware because they start and end with nonzero velocities. A simple workaround is to repeat the first and last frames so that the reference is static at the beginning and end of the execution. This edit introduces velocity discontinuities at the transitions between standing and jumping, which TO can smooth out. We apply this edit to all hardware experiments. The left subfigure of \Fig{\ref{fig:edit_raw_kto_kdto}} shows the learning curves for this edit applied to the baseline case where the jump segment itself remains unmodified. KTO improves the convergence speed, and the gap between KTO and KDTO is small. All three variants reach similar final tracking errors, suggesting that this edit introduces only minor dynamic artifacts in our setting.

\subsection{Highly Dynamic Motion Tracking}
\label{sec:dyn_motion_track}
\begin{figure}
    \centering
    \includegraphics[width=0.9\linewidth]{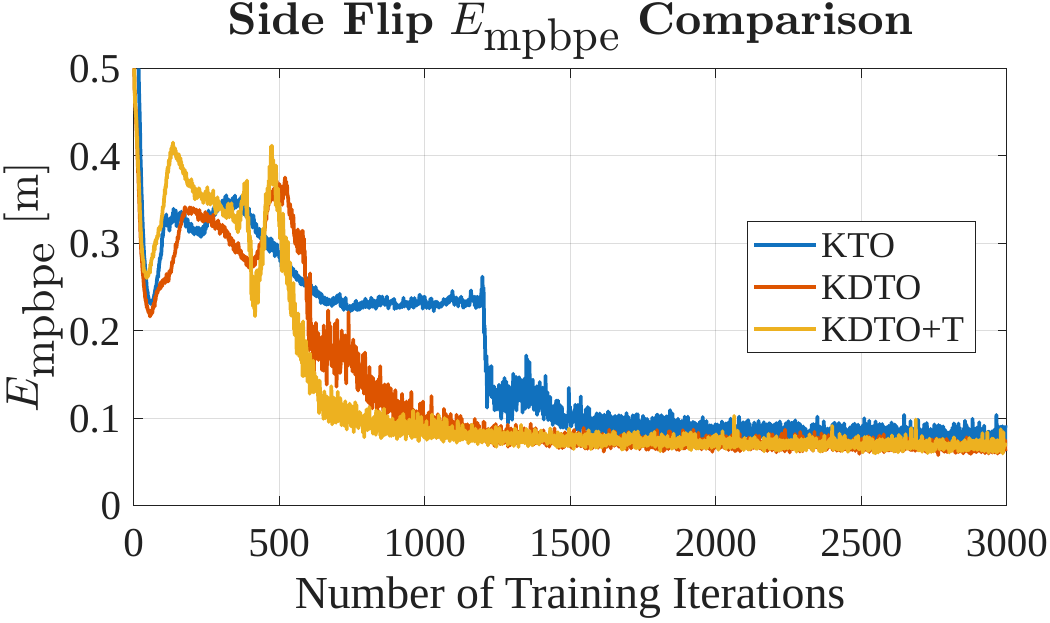}
    \caption{Comparison of RL $E_\mpbpe$ learning curves of side flip tracking when the reference motions are provided by KTO, KDTO, KDTO+T.}
    \label{fig:dyn_feas}
\end{figure}

The advantage of recovering dynamically feasible trajectories with KDTO is most pronounced for highly dynamic motions, such as the side flip motion shown in \Fig{\ref{fig:side_flip_hardware}}. Beyond producing a kinematically consistent trajectory, KDTO also yields joint torque references that can accelerate policy learning. In \Fig{\ref{fig:dyn_feas}} we compare $E_\mpbpe$ learning curves for side-flip tracking under three reference variants. The KTO and KDTO runs use the standard BeyondMimic~\cite{liao2025beyondmimic} reward, while KDTO+T adds an exponential joint torque tracking term $r_{\btau_j}$ tracking the optimal KDTO joint torques $\btau_{j,\KDTO}^*$, which is defined as
\begin{equation}
    \begin{aligned}
        r_{\btau_j} &= \exp\left(
        \tau_{j, \text{MSE}} / \sigma
        \right) \\
        \tau_{j, \text{MSE}} &= 
        \frac{1}{N_j}\LtwoNormSquare{
            \left( \btau_j - \btau_{j,\KDTO}^* \right)
            ./ \btau_{j,\max}
        }.
    \end{aligned}
    \label{eq:torq_track}
\end{equation}
This term encourages joint torques $\btau_j$ to track KDTO-optimized torques $\btau_{j,\KDTO}$. Here, $\tau_{j,\text{MSE}}$ is the mean-squared normalized torque error, with $./$ denoting element-wise division. $N_j$ is the number of joints and $\btau_{j,\max}$ is the torque limit. As shown in \Fig{\ref{fig:dyn_feas}}, the KTO learning curve plateaus for roughly 500 iterations before converging, as training struggles near the most challenging flip timestep when the robot is inverted. In this regime, KTO must simultaneously learn to track the difficult motion while correcting dynamics infeasibility. By contrast, KDTO exhibits a much shorter plateau because the reference motion is already dynamically feasible, and KDTO+T converges even faster thanks to the additional guidance from the torque-tracking reward.

\section{Conclusion}
In this paper, we presented a pipeline for transferring task space human motion to humanoid robots with no per-motion tuning. We first introduced a URDF calibration procedure that aligns human skeletal dimensions with the target robot prior to reference generation, producing task-space targets that are kinematically consistent with the robot morphology. This structural alignment substantially reduces the need for manual IK weight adjustment and orientation-offset tuning.

We then investigated the role of dynamics consistency in tracking edited and highly dynamic motions and demonstrated that KDTO recovers dynamically feasible trajectories and provides additional joint torque references, which improve both learning efficiency and tracking performance.


\balance

\bibliographystyle{IEEEtran}
\bibliography{references/model_based, references/rl, references/robot, references/retarget}
\end{document}